\documentclass[sigconf]{acmart}
\usepackage{graphicx}
\usepackage{amsmath}
\usepackage{graphicx}
\usepackage{mathtools}

\usepackage{amssymb}
\usepackage{amsthm}
\usepackage{booktabs}
\usepackage{algorithm}
\usepackage{algorithmic}
\usepackage{subfig} 
\usepackage{multirow}
\AtBeginDocument{%
  \providecommand\BibTeX{{%
    \normalfont B\kern-0.5em{\scshape i\kern-0.25em b}\kern-0.8em\TeX}}}

\setcopyright{acmcopyright}
\copyrightyear{2021}
\acmYear{2021}
\acmDOI{x.xx/xxxx.xxxx}

\acmConference[-]{-}{-}{-}
\acmPrice{-}
\acmISBN{-}

\begin{document}

\title{Unsupervised Domain Adaptation for Semantic Segmentation via Low-level Edge Information Transfer}

\author{Hongruixuan Chen}
\affiliation{%
  \institution{Wuhan University}
  \city{Wuhan}
  \country{China}
}
\email{Qschrx@whu.edu.cn}

\author{Chen Wu}
\affiliation{%
  \institution{Wuhan University}
  \city{Wuhan}
  \country{China}
}
\email{chen.wu@whu.edu.cn}

\author{Yonghao Xu}
\affiliation{%
  \institution{Wuhan University}
  \city{Wuhan}
  \country{China}
}
\email{yonghaoxu@ieee.org}

\author{Bo Du}
\affiliation{%
  \institution{Wuhan University}
  \city{Wuhan}
  \country{China}
}
\email{gunspace@163.com}

% \author{Jian Ye}
% \affiliation{%
%   \institution{Wuhan University}
%   \city{Wuhan}
%   \state{Hubei}
%   \country{China}
% }
% \email{bodu@whu.edu.cn}


\begin{abstract}
    Unsupervised domain adaptation for semantic segmentation aims to make models trained on synthetic data (source domain) adapt to real images (target domain). Previous feature-level adversarial learning methods only consider adapting models on the high-level semantic features. However, the large domain gap between source and target domains in the high-level semantic features makes accurate adaptation difficult. In this paper, we present the first attempt at explicitly using low-level edge information, which has a small inter-domain gap, to guide the transfer of semantic information. To this end, a semantic-edge domain adaptation architecture is proposed, which uses an independent edge stream to process edge information, thereby generating high-quality semantic boundaries over the target domain. Then, an edge consistency loss is presented to align target semantic predictions with produced semantic boundaries. Moreover, we further propose two entropy reweighting methods for semantic adversarial learning and self-supervised learning, respectively, which can further enhance the adaptation performance of our architecture. Comprehensive experiments on two UDA benchmark datasets demonstrate the superiority of our architecture compared with state-of-the-art methods. 
\end{abstract}



\maketitle

\section{Introduction}
\par Semantic segmentation is a fundamental task in image processing, which aims to assign semantic labels to all pixels in a given image. Obtaining precise semantic segmentation results is significant for many vision-based applications \cite{zhang2016instance,wang2016temporal,xu2020march,pham2000current,Wu2021,Chen2019a}. Nowadays, deep learning-based models, especially convolutional neural networks (CNNs) \cite{lecun1998gradient,krizhevsky2012imagenet}, have achieved promising progress in semantic segmentation. To train a good segmentation network, a large number of fully annotated images are often required. Nevertheless, collecting large-scale datasets with accurate pixel-level annotation is time-consuming \cite{cordts2016cityscapes}. To reduce labeling consumption, an alternative way is utilizing synthetic images with precise pixel-level annotations to train the deep models. These synthetic images and corresponding annotations can be automatically generated by game engines, such as Grand Theft Auto V (GTAV) \cite{richter2016playing}. However, due to the large domain gap caused by the appearance difference between synthetic images and real images, the models trained on the synthetic images (source domain) inevitably face severe performance degradation on the real-world image datasets (target domain). 

\begin{figure}[!t]
    \centering
    \includegraphics[scale=0.3]{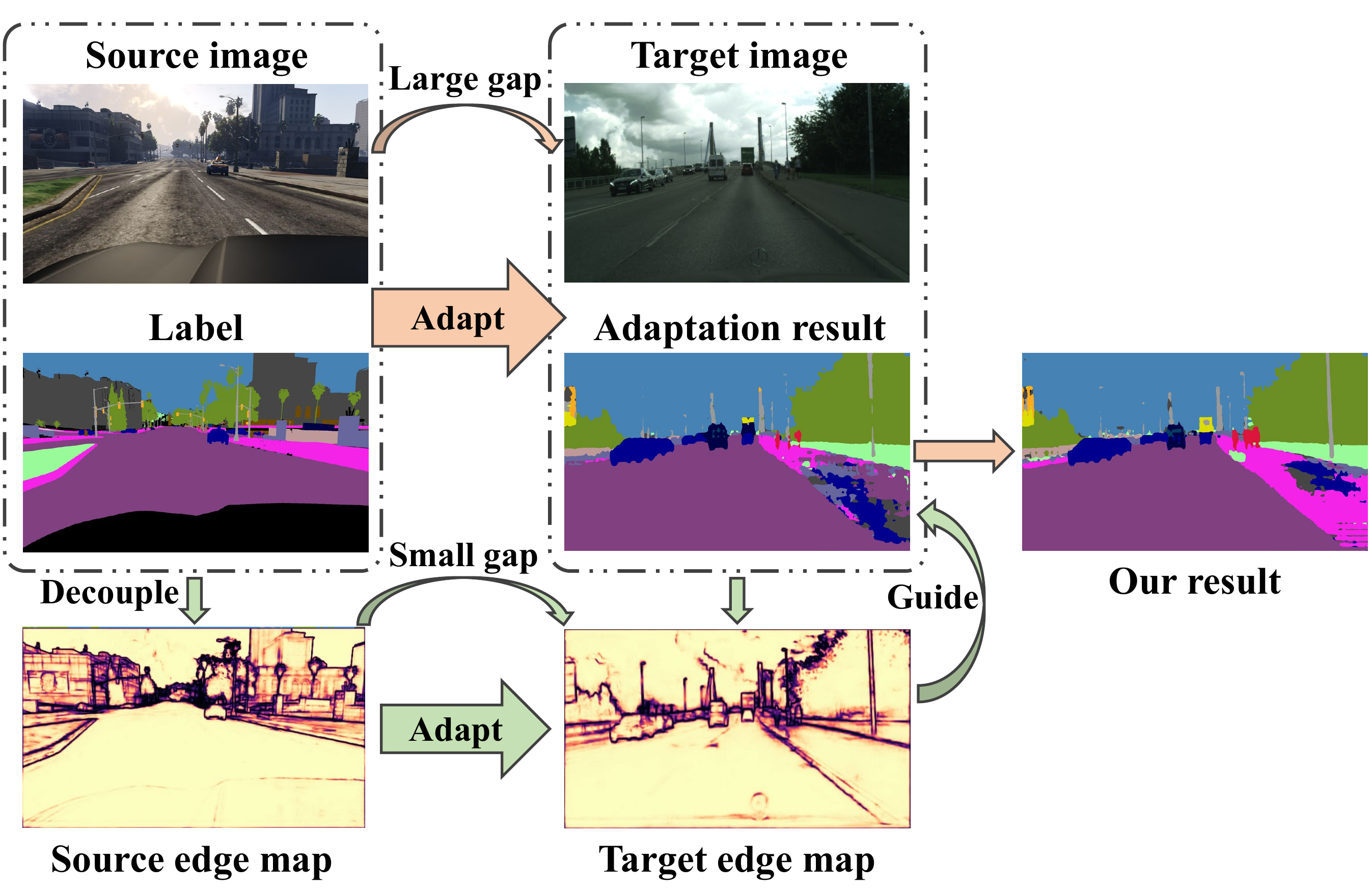}
    \caption{We propose to explicitly use low-level edge information with a small inter-domain gap for UDA in semantic segmentation. Compared with high-level semantic information, low-level edge information is easier to adapt; thus high-quality edge maps can be produced over the target domain. Since the edge information can reflect the boundaries of semantic category, it can be used to guide the transfer of semantic information.}
    \label{fig:1}
\end{figure}

\par To address this issue, unsupervised domain adaptation (UDA) methods have been introduced to reduce the domain gap between labeled source domain and unlabeled target domain. In terms of the semantic segmentation task, adversarial learning-based UDA approaches demonstrate good efficiency in aligning domain gaps in the feature-level \cite{hoffman2016fcns,tsai2018learning,vu2019advent,luo2019taking,yu2021dast}. All of these methods align high-level feature distributions of different domains since high-level features contain abundant semantic category information. However, the large inter-domain gap in the high-level semantic representations makes the accurate alignment difficult. As pointed by Luo \emph{et al.} \cite{luo2019taking}, directly aligning the high-level semantic features may lead to negative transfer and damage the adaptation performance in the originally well-aligned regions. To address this issue, they propose a local score alignment map to guide the transfer of semantic information. 

\par In this paper, we provide a different viewpoint for addressing this issue. As argued in \cite{long2015learning}, in contrast to deep feature representations with large domain gaps and poor transferability, feature representations extracted by shallow convolutional layers are often general. Then, according to the visualization results of CNN reported in \cite{zeiler2014visualizing}, the feature representations extracted by CNN show strong hierarchical nature and the shallow layers highly respond to low-level edge and color information. Based on these arguments and observations, we argue that low-level edge feature representations have a smaller inter-domain gap in comparison with high-level semantic features. Intuitively, it could be also observed that although the synthetic image and real-world image are quite different in appearance, the object shapes of the same category are very similar. Moreover, there exists a strong interaction between the edge information and the semantic information: the edge information can reflect the boundaries of semantic category. 
% , and the effect of low-level features is often ignored. In this paper, we argue that low-level edge feature representations have a small domain gap and can be explicitly utilized for cross-domain semantic segmentation: 1) Although the synthetic image and real-world image are quite different in color and texture, the object shapes of the same category are very similar; 2) Moreover, there exists a strong interaction between the edge information (low-level) and the semantic information (high-level): the edge information can reflect the category boundaries, which can be utilized to guide the semantic segmentation results. Hence, as shown in Figure \ref{fig:1}, we can treat low-level edge information as the transferrable factor that could be used to facilitate transfer for high-level semantic segmentation, thereby obtaining more accurate results. 

\par Consequently, as shown in Figure \ref{fig:1}, we treat low-level edge information as the transferable factor that could be used to facilitate transfer for high-level semantic information. Specifically, we present a semantic-edge domain adaptation (SEDA) architecture consisting of a semantic stream and an edge stream. In our architecture, the edge information is decoupled from the mainstream semantic network and is explicitly processed by an independent stream. The semantic stream adopts the existing entropy adversarial learning method as the basis. To better adapt hard target images, we present an entropy reweighting method to assign larger weights to harder images. For edge stream, we train it with the source semantic boundaries and adapt the source and target edge features through adversarial learning. An edge consistency loss function is applied to encourage the semantic segmentation predictions to correctly align with the semantic boundaries. As the target results with more accurate boundaries are obtained, we use self-supervised learning (SL) to further fit the distribution of target domain. Furthermore, to overcome the issues of standard self-supervised learning, an uncertainty-adaptive self-supervised learning (UASL) is presented. The contributions of our work can be concluded as follows:
\begin{enumerate}
\item This paper proposes a semantic-edge domain adaptation architecture, which presents the first attempt at explicitly using edge information that has a small inter-domain gap for facilitating the transfer of high-level semantic information.
\item Two entropy-based reweighting methods are proposed to improve adversarial learning and self-supervised learning, enabling our architecture to learn better domain-invariant representations.
\item Experiments on two challenging benchmark adaptation tasks demonstrate that the proposed method can obtain better results than existing state-of-the-art methods.
\end{enumerate}
\section{Related Work}
\subsection{Semantic Segmentation}
\par Semantic segmentation is one of the most challenging computer vision tasks, which aims to predict pixel-level semantic labels for a given image. Following the work in \cite{long2015fully}, it has become mainstream to use fully convolutional network (FCN) architecture for tackling the semantic segmentation task, and many effective models have been presented \cite{wang2016objectness,badrinarayanan2017segnet,chen2017deeplab,zhao2017pyramid,xu2020march,gu2020context,liu2020dynamic}. Besides, some probability graph models like conditional random filed \cite{koltun2011efficient} are used as an effective post-processing method for improving performance. More recently, some work introduces multi-task learning into semantic segmentation \cite{dai2016instance,chen2016semantic,kokkinos2017ubernet,takikawa2019gated}, which combines networks for complementary tasks to improve semantic segmentation accuracy. Nevertheless, to train these semantic segmentation models, numerous real-world images with pixel-level annotations are required, which are usually difficult to collect. An alternative way is to train these models with photo-realistic synthetic data.

\subsection{UDA for Semantic Segmentation}
\par Unsupervised domain adaptation aims to align the domain distribution shift between labeled source data and unlabeled target data \cite{zou2020unsupervised,zhuo2017deep,li2019joint,jiang2020resource,chen2020dsdanet}. A very attractive application of UDA is using photo-realistic synthetic data to train semantic segmentation models, and a variety of methods have been presented, which can be broadly divided into three types: adversarial learning based approach \cite{hoffman2016fcns,tsai2018learning,tsai2019domain,vu2019advent,li2019bidirectional, kim2020learning,wang2020classes}, image translation approach \cite{hoffman2018cycada,zhang2018fully,li2019bidirectional,choi2019self,yang2020fda}, and self-supervised learning approach \cite{xu2019self, choi2019self,song2020learning,zou2018unsupervised,zhang2019curriculum,lian2019constructing}. 

\par \textbf{Adversarial learning} methods involve two networks. A generative network predicts the segmentation maps for the input source or target images. Another discriminator network takes the feature maps from generative network and tries to predict the domain type of feature maps, while generative network tries to fool the discriminator. By iteratively repeating this process, the two domains would have a similar distribution. In \cite{hoffman2016fcns}, the adversarial approach for first applied to UDA for semantic segmentation. In  \cite{vu2019advent}, adversarial learning is performed to match the entropy maps of two domains. In \cite{luo2019taking}, a local alignment score map is designed to evaluate the category-level alignment degree for guiding the transfer of semantic features. In \cite{yu2021dast}, an attention-based discriminator network is presented to adaptively measure the hard-adapted semantic features. 

\par \textbf{Image translation} methods directly apply adversarial models or style transfer approaches to transform the source images into target-style images for aligning the domain gap. In \cite{hoffman2018cycada}, CycleGAN \cite{zhu2017unpaired} is used to transform the synthetic images of the source domain to the style of the target images. In \cite{zhang2018fully}, a style transfer network is presented to make the images from two domains visually similar. Recently, Yang \emph{et al.} \cite{yang2020fda} use fast Fourier transform to reduce the appearance difference between images from different domains.   

\begin{figure*}[ht]
    \centering
    \includegraphics[scale=0.68]{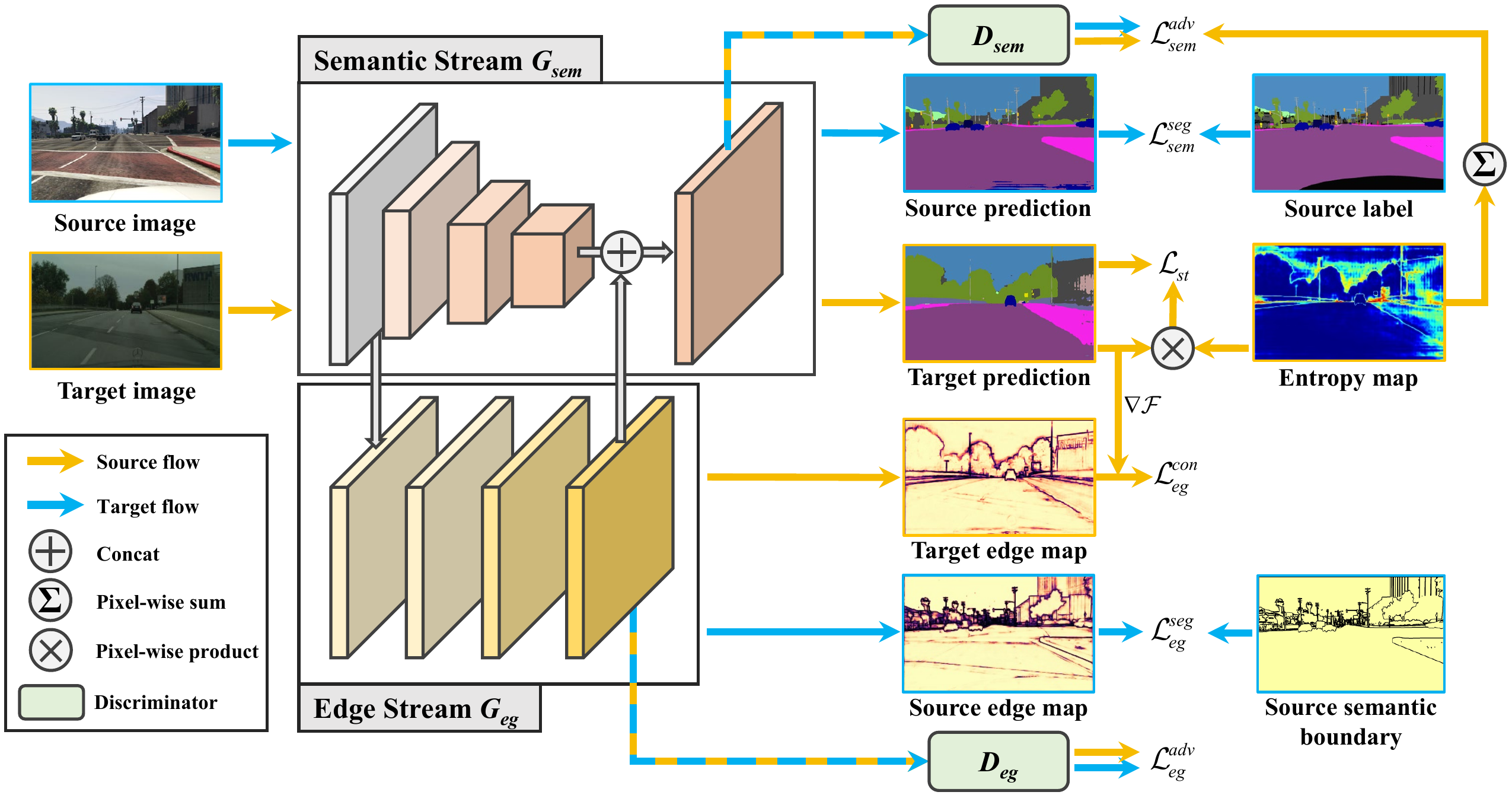}
    \caption{The overall architecture of our SEDA architecture, which composed of three parts: 1) In semantic information transfer, the feature-level adversarial learning approach is applied to align the semantic distributions of source and target domains. Besides, the entropy map of target output is utilized to weight each target sample at the image-level; 2) In edge information transfer, the edge information is decoupled from $G_{sem}$ and independently processed by $G_{eg}$. Feature-level adversarial learning is also performed to align edge feature distributions of two domains. Then, the target semantic boundary map is used to guide the target semantic segmentation; 3) In uncertainty-adaptive self-supervised learning, the target prediction is weighted by entropy map at pixel-level and treated as pseudo-labels to train $G_{sem}$. }
    \label{fig:2}
\end{figure*}

\par \textbf{Self-supervised learning} is another effective UDA approach. In the field of UDA for semantic segmentation, self-supervised learning methods use the target prediction as pseudo-labels to train the segmentation network, which could make the model implicitly learn the domain-invariant representations \cite{zou2018unsupervised,yu2021dast}. In \cite{zou2018unsupervised}, a class balancing strategy and spatial prior are presented to guide the self-supervised learning in target domain. In \cite{lian2019constructing}, a pyramid curriculum strategy based on multi-scale pooling is proposed to select reliable pseudo-labels to train the segmentation network. Moreover, self-supervised learning could be combined with adversarial learning or image translation to further boost UDA performance \cite{yang2020fda,kim2020learning,pan2020unsupervised,yu2021dast}. 

% Selecting an appropriate threshold for generating pseudo-labels is the key for self-supervised learning. 

\par In this paper, unlike the previous adversarial learning methods that only focus on aligning the high-level semantic features, the proposed method simultaneously aligns semantic and edge features and utilizes adapted edge features to facilitate the transfer of semantic features. Moreover, we further present an entropy reweighting semantic adversarial learning strategy and an uncertainty-adaptive self-supervised learning approach to enhance the UDA performance.

\section{Methodology}
\par In UDA for semantic segmentation, the labeled source domain is denoted as $\mathcal{D}_{\mathcal{S}}=\left\{\left(X_{s}, Y_{s}\right)\right\}_{s \in \mathcal{S}}$, and the unlabeled target domain is denoted as $\mathcal{D}_{\mathcal{T}}=\left\{X_{t}\right\}_{t \in \mathcal{T}}$, where $X_{s} \in \mathbb{R}^{H \times W \times 3}$ is a source image, $Y_{s} \in \mathbb{R}^{H \times W \times C}$ is the one-hot semantic label associated with $X_{s}$, and $X_{t} \in \mathbb{R}^{H \times W \times 3}$ is a target image. Our goal is to utilize the low-level edge information, which is relatively easier to be transferred on the two domains, to guide the semantic segmentation over the target domain, thereby obtaining desirable prediction performance. Figure \ref{fig:2} illustrates our architecture that mainly contains three parts: semantic information transfer, edge information transfer, and uncertainty-adaptive self-supervised learning.

\subsection{Semantic Information Transfer}
\par Semantic stream $G_{sem}$ is the basis of our architecture. Specifically, we use entropy adversarial learning method as our baseline, which has yielded promising results in UDA for semantic segmentation \cite{vu2019advent} and has served as the basis of more advanced methods \cite{vu2019dada,pan2020unsupervised}. First, $G_{sem}$ is trained by minimizing cross-entropy loss $\mathcal{L}_{sem}^{seg}$ over source data:
\begin{equation}
    \mathcal{L}_{sem}^{seg}=-\sum_{h, w} \sum_{c} Y_{s}^{(h, w, c)} \log P_{s}^{(h, w, c)}
\label{eq:0}
\end{equation}
where $P_{s}\in \mathbb{R}^{H \times W \times C}$ is the source semantic prediction map generate by $G_{sem}$. Besides, to overcome the negative effect of class imbalance problem, lov{\'a}sz-softmax loss \cite{berman2018lovasz} is imposed on source data.

\par Subsequently, $G_{sem}$ takes a target image as input and output the semantic prediction map $P_{t}$. Then, the weighted self-information map $I_{t} \in \mathbb{R}^{H \times W \times C}$ is calculated:
\begin{equation}
    I_{t}^{(h, w)}=-P_{t}^{(h, w)} \log P_{t}^{(h, w)}
\label{eq:1}
\end{equation}

\par To reduce the domain gap, in entropy adversarial learning, a discriminator $D_{sem}$ is trained to predict the domain type for the weighted self-information map, and $G_{sem}$ is trained to fool $D_{sem}$. However, the entropy adversarial learning treats all the target images equally, but there exist easy-adapt images with simple scenes and hard-adapt images with difficult scenes in the target domain. To better optimize these hard-adapt samples, we argue that harder samples need to contribute more loss during the training stage. Since the entropy map can reflect the confidence levels of the target predictions \cite{vu2019advent}, we utilize the pixel-wise sum of target entropy map to measure the difficulty for each target image: 
\begin{equation}
    \left\{
        \begin{array}{l}
            E_{t}^{(h, w)}=-\frac{1}{\log C} \sum\limits_{c} P_{t}^{(h, w, c)} \log P_{t}^{(h, w, c)} \\
            \mathcal{E}_{t}=\sum\limits_{h, w} E_{t}^{(h, w)}
        \end{array}
    \right.
   \label{eq:1.5}
\end{equation}
where $\mathcal{E}_{t}$ is the pixel-wise sum of entropy map $E_{t}$. If a target image has a low overall entropy value, it can be regarded as an easy-adapt sample, otherwise it is a hard-adapt sample. Based on this assumption, we propose an entropy reweighting adversarial loss:
\begin{equation}
    \begin{split}
        \mathcal{L}_{sem}^{a d v}=&-\sum_{h, w} \log \left(1-D_{sem}\left(I_{s}\right)\right) \\
        &-\left(1+\left(\alpha \mathcal{E}_{t}\right)^{2} \right)\sum_{h, w} \log \left(D_{sem}\left(I_{t}\right)\right)
    \end{split}
\label{eq:2}
\end{equation}
where $I_{s}$ is the source weighted self-information map, and $\alpha$ is a weight factor. Noteworthy, we adopt the square of entropy to enlarge the loss difference between easy-adapt samples and hard-adapt samples. 

\par Through optimizing $\mathcal{L}_{s e m}^{seg}$ and $\mathcal{L}_{s e m}^{a d v}$, the two domains are aligned at the semantic feature maps to some extent. Next, we explicitly use the low-level edge information to further facilitate the transfer of semantic features.

\subsection{Edge Information Transfer}
\par In previous methods, edge information is entangled with other types of information in the segmentation network and is implicitly adapted through adversarial learning, which makes it difficult to use edge information for facilitating the transfer of semantic information.

\par To explicitly use low-level edge information, we use an independent edge stream $G_{eg}$ to decouple edge information from $G_{sem}$. In terms of the specific network architecture, we adopt a lightweight auxiliary network introduced in \cite{takikawa2019gated} for edge detection. A gated convolutional layer is introduced in $G_{eg}$ to ensure that $G_{eg}$ only processes edge-relevant information. Specifically, $G_{eg}$ takes the output of the first convolutional layer of $G_{sem}$ as input and aims to yield precise semantic boundary maps $\mathcal{B}_{s}$ and $\mathcal{B}_{t}$. To this end, $G_{eg}$ is first trained by minimizing binary cross-entropy loss $\mathcal{L}_{eg}^{seg}$ over the source domain. Ground truth of semantic boundaries can be directly generated from source semantic labels.  

\par Through optimizing $\mathcal{L}_{eg}^{seg}$, $G_{eg}$ is capable of generating precise semantic boundaries for source domain data. Compared to high-level semantic features, the low-level edge features have a smaller domain gap. Moreover, due to optimizing $\mathcal{L}_{s e m}^{a d v}$, the inter-domain gap in the features of shallow layers is further reduced. Accordingly, despite merely supervised by source edge information, $G_{eg}$ can generate decent boundaries in the target domain. To produce higher quality semantic boundaries for target domain, similar to semantic stream, we also introduce a discriminator $D_{eg}$ for predicting the domain labels for the edge feature maps, while $G_{eg}$ is trained to fool $D_{eg}$:
\begin{equation}
    \mathcal{L}_{e g}^{a d v}=-\sum_{h, w}\left(\log \left(1-D_{eg}\left(\mathcal{H}_{s}\right)\right)+\log \left(D_{eg}\left(\mathcal{H}_{t}\right)\right)\right)
\label{eq:3}
\end{equation}
where $\mathcal{H}_{s}$ and $\mathcal{H}_{t}$ are the edge features of source and target domains from the last layer of $G_{eg}$. 

\par Subsequently, we add the target edge feature map back to $G_{sem}$, thereby guiding the semantic segmentation over the target domain. However, this way can only implicitly refine the semantic segmentation results, which cannot guarantee consistency between the boundary map and the predicted semantic segmentation map. To explicitly encourage the target semantic segmentation maps to align with the boundary map, we introduce an edge consistency loss $\mathcal{L}_{e g}^{c o n}$:
\begin{equation}
    \left\{\begin{array}{l}
        \mathcal{P}_{t}=\frac{1}{\sqrt{2}}\left\|\nabla\left(\mathcal{F} * \arg \max\limits_{c} P_{t}\right)\right\| \\
        \mathcal{L}_{e g}^{c o n}=\sum\limits_{\mathcal{N}^{+}}\left|\mathcal{P}_{t}^{(\mathcal{N}^{+})}-\mathcal{B}_{t}^{(\mathcal{N}^{+})}\right|
        \end{array}\right.
\label{eq:4}
\end{equation}
where $\mathcal{P}_{t}$ is the semantic boundary map computed by taking a spatial derivative on the target segmentation output, $\mathcal{F}$ is Gaussian filter, and $\mathcal{N}^{+}$ contains coordinates of all boundary pixels in both $\mathcal{P}_{t}$ and $\mathcal{B}_{t}$. 

\par $\mathcal{L}_{e g}^{c o n}$ aims at ensuring that target semantic boundary pixels are penalized if there is a mismatch with boundaries predicted by $G_{eg}$. By optimizing $\mathcal{L}_{e g}^{c o n}$, $G_{sem}$ can be guided by the target boundary map, thereby generating more accurate target prediction map. Furthermore, since argmax operator is not differentiable, the Gumbel softmax trick \cite{jang2016categorical} is adopted to approximate the partial derivatives of $\mathcal{L}_{e g}^{c o n}$ to a given parameter during the backward propagation stage. 

\subsection{Uncertainty-Adaptive Self-Supervised Learning}
\par By means of using the low-level edge information to guide the transfer of semantic information, $G_{sem}$ can generate more accurate semantic segmentation results over the target domain. Subsequently, considering the complex distribution of real-world data (target domain), we apply the self-supervised learning strategy to make our architecture further fit the distribution of the target domain. 
\par The standard self-supervised learning method \cite{zou2018unsupervised,pan2020unsupervised} sets a threshold to select high-confident target pseudo-labels. However, it is difficult to choose a suitable threshold: an over-large threshold could make the available target information very less, and an over-small threshold could produce too many noisy labels, damaging the adaptation performance. Besides, the confidence levels of these selected pseudo-labels are also different. To address these issues, we present an uncertainty-adaptive self-supervised loss that adopts entropy to adaptively estimate uncertainty and reweight target prediction at pixel-level:
\begin{equation}
    \mathcal{L}_{uasl}=-\sum_{h, w}\left(1-E_{t}^{(h, w)}\right)^{2} \sum_{c} \hat{Y_{t}}^{(h, w, c)} \log P_{t}^{(h, w, c)}
\label{eq:5}
\end{equation}
where $\hat{Y_{t}}$ is the one-hot semantic pseudo-labels.

\begin{table*}[ht]
    \caption{Evaluation results of semantic segmentation by adapting from GTAV to Cityscapes. The mechanism “T”, “A”, and “S” mean image translation, adversarial training, and self-supervised learning, respectively. The best results are highlighted in \textbf{bold}.}
    \setlength\tabcolsep{2.3pt}
    \centering
    \begin{tabular}{l | c | c c c c c c c c c c c c c c c c c c c | c}
    \toprule
    \multicolumn{22}{c}{GTAV$\rightarrow$Cityscapes}\\
    \hline
    Methods & \rotatebox{90}{Mech.} & \rotatebox{90}{road} & \rotatebox{90}{sidewalk } & \rotatebox{90}{building} & \rotatebox{90}{wall} & \rotatebox{90}{fence} & \rotatebox{90}{pole} & \rotatebox{90}{light} & \rotatebox{90}{sign} & \rotatebox{90}{veg.} & \rotatebox{90}{terrain} & \rotatebox{90}{sky} & \rotatebox{90}{person} & \rotatebox{90}{rider} & \rotatebox{90}{car} & \rotatebox{90}{truck} & \rotatebox{90}{bus} & \rotatebox{90}{train} & \rotatebox{90}{mbike} & \rotatebox{90}{bike} & mIoU \\
    \hline
    AdaSegNet \cite{tsai2018learning}& A	& 86.5&	36.0&	79.9&	23.4	&23.3&	23.9&	35.2&	14.8&	83.4&	33.3	&75.6&	58.5	&27.6	&73.7&	32.5&	35.4&	3.9	&30.1	& 28.1 &42.4 \\
    ADVENT  \cite{vu2019advent} & A &	89.4&	33.1&	81.0&	26.6&	26.8&	27.2&	33.5&	24.7&	83.9&	36.7&	78.8&	58.7&	30.5&	84.8&	38.5&	44.5&	1.7&	31.6&	32.4&	45.5 \\
    CLAN \cite{luo2019taking}& A &	87.0&	27.1&	79.6&	27.3&	23.3&	28.3&	35.5&	24.2&	83.6&	27.4&	74.2&	58.6&	28.0&	76.2&	33.1&	36.7&	6.7&	31.9&	31.4&	43.2\\
    AdaptPatch \cite{tsai2019domain}& A &	92.3&	51.9&	82.1&	29.2&	25.1	&24.5&	33.8&	33.0&82.4&	32.8&	82.2&	58.6&	27.2&	84.3&	33.4&	46.3&	2.2&	29.5&	32.3&	46.5\\
    % SIBAN \cite{luo2019significance}& A &	88.5&	35.4&	79.5&	26.3&	24.3&	28.5&	32.5&	18.3&	81.2&	40.0&	76.5&	58.1&	25.8&	82.6&	30.3&	34.4&	3.4&	21.6&	21.5&	42.6\\
    PyCDA \cite{lian2019constructing}&S& 90.5& 36.3& 84.4 &32.4 &28.7 &34.6 &36.4 &31.5 &86.8 &37.9 &78.5& 62.3& 21.5& 85.6 &27.9 &34.8 &\textbf{18.0} &22.9& 49.3& 47.4\\
    CCM \cite{li2020content} &S& 93.5 & 57.6 & 84.6 & 39.3 & 24.1 & 25.2 & 35.0 & 17.3& 85.0 & 40.6&  86.5&  58.7 & 28.7 & 85.8 & \textbf{49.0}&  \textbf{56.4} & 5.4 & 31.9 & 43.2&  49.9 \\
    \hline
    FDA	\cite{yang2020fda}& TS & 92.5&	53.3&	82.4&	26.5&	27.6&	36.4&	40.6&	38.9&	82.3&	39.8&	78.0&	62.6&	34.4&	84.9&	34.1&	53.1&	16.9&	27.7&	46.4&	50.4\\
    IntraDA \cite{pan2020unsupervised}& AS &	90.6&	37.1&	82.6&	30.1&	19.1&	29.5&	32.4&	20.6&	85.7&	40.5&	79.7&	58.7&	31.1&	86.3&	31.5&	48.3&	0.0&	30.2&	35.8&	46.3\\
    FADA \cite{wang2020classes}& AS &	91.0&	50.6&	\textbf{86.0}&	\textbf{43.4}&	29.8&	\textbf{36.8}&	\textbf{43.4}&	25.0&	86.8&	38.3&	\textbf{87.4}&	\textbf{64.4}&	\textbf{38.0}&	85.2&	31.6&	46.1&	6.5&	25.4&	37.1&	50.1\\
    % CCM \cite{li2020content}& S & 93.5&	57.6&	84.6&	39.3&	24.1&	25.2&	35.0&	17.3&	85.0&	40.6&	86.5&	58.7&	28.7&	85.8&	\textbf{49.0}&	\textbf{56.4}&	5.4&	31.9&	43.2&	49.9\\
    DAST \cite{yu2021dast} & AS & 92.2&	49.0&	84.3&	36.5&	28.9&	33.9&	38.8&	28.4&	84.9&	41.6&	83.2&	60.0&	28.7&	87.2&	45.0&	45.3&	7.4&	33.8&	32.8&	49.6\\
    \hline
    BDL \cite{li2019bidirectional} & TAS& 91.0 &44.7& 84.2& 34.6& 27.6 &30.2 &36.0 &36.0 &85.0 &43.6 &83.0 &58.6 &31.6 &83.3 &35.3 &49.7 &3.3 &28.8 &35.6 &48.5 \\
    TIR	\cite{kim2020learning}& TAS & 92.9&	55.0&	85.3&	34.2&	31.1&	34.9&	40.7&	34.0&	85.2&	40.1&	87.1&	61.0&	31.1&	82.5&	32.3&	42.9&	0.3&	36.4&	46.1&	50.2\\
    LDR \cite{yang2020label} & TAS &90.8 &41.4 &84.7 &35.1 &27.5 &31.2 &38.0 &32.8& 85.6 &42.1& 84.9 &59.6 &34.4 &85.0 &42.8 &52.7& 3.4& 30.9& 38.1& 49.5 \\ 
    UDACT \cite{lee2020unsupervised} & TAS & \textbf{95.3}&	\textbf{65.1}&	84.6&	33.2&	23.7&	32.8&	32.7&	36.9&	86.0&	41.0&	85.6&	56.1&	25.9&	86.3&	34.5&	39.1&	11.5&	28.3&	43.0&	49.6\\
    \hline
    % SourceOnly \cite{tsai2019domain}& -	& 75.8	& 16.8	& 77.2	& 12.5	& 21.0	& 25.5	& 30.1	& 20.1	& 81.3	& 24.6	& 70.3	& 53.8 & 26.4 & 49.9 & 17.2	& 25.9	& 6.5 & 25.3 & 36.0 & 36.6 \\	
    SourceOnly & -	& 64.6	& 27.3	& 76.9	& 19.1	& 21.1	& 27.0	& 32.1	& 18.5	& 81.2	& 14.5	& 72.4	& 55.4 & 21.6 & 62.9 & 29.4	& 8.4 & 2.4 & 24.2 & 35.0 & 36.5 \\	
    % Baseline& A &	88.5&	32.6&	81.5&	27.3&	23.3&	30.7&	33.2&	26.9&	82.1&	30.6&	77.7&	56.7&	25.4&	84.9&	31.0&	42.3&	0.5&	25.4&	32.5&	43.9\\ 
    % Ours w/o UAST & A &	91.9&	50.5&	83.1&	27.9&	\textbf{26.3}&	34.8&	38.6&	\textbf{38.8}&	84.2&	28.1&	81.6&	62.5&	\textbf{31.1}&	\textbf{87.7}&	37.0&	41.8&	2.6&	33.1&	45.2&	49.2\\ 
    % Ours w/ UAST & AS &	93.3&	61.1&	85.0&	33.1&	\textbf{30.7}&	32.0&	39.0&	\textbf{47.3}&	86.4&	28.1&	82.6&	63.2&	\textbf{38.4}&	\textbf{87.4}&	45.8&	42.6&	0.0&	\textbf{39.5}&	\textbf{52.3}&	\textbf{52.0}\\ 
    Ours  & AS &	94.0&	61.8&	85.8&	29.2&	\textbf{32.5}&	35.4&	40.6&	\textbf{43.3}&	\textbf{87.2}&	\textbf{43.9}&	84.4&	63.8&	29.1&	\textbf{88.7}&	46.0&	49.9&	0.0&	\textbf{43.7}&	\textbf{49.9}&	\textbf{52.8}\\ 
    \toprule
    \end{tabular}
    \label{tab:1}
\end{table*}

\par Both $\mathcal{L}_{sem}^{adv}$ and $\mathcal{L}_{uasl}$ adopt entropy to measure uncertainty and reweight samples. However, $\mathcal{L}_{sem}^{adv}$ uses the sum of entropy to weight the target image at the image-level, but $\mathcal{L}_{s t}$ uses entropy map to weight target prediction at the pixel-level. In $\mathcal{L}_{sem}^{adv}$, the image with larger entropy means harder to adapt and is assigned more weight. In contrast, the pixel with a smaller entropy value represents higher confidence and is more highlighted in $\mathcal{L}_{uasl}$.

\par Finally, our complete loss function $\mathcal{L}$ is formed by all the loss functions: 
\begin{equation}
\begin{aligned} 
\mathcal{L} &=\mathcal{L}_{s e m}^{s e g}+\lambda_{1} \mathcal{L}_{s e m}^{a d v}+\lambda_{2} \mathcal{L}_{e g}^{s e g} +\lambda_{3} \mathcal{L}_{e g}^{a d v}\\ &+\mathcal{L}_{e g}^{c o n}+\mathcal{L}_{u a s l} \end{aligned}    
\label{eq:6}
\end{equation}
where $\lambda_{1}$ to $\lambda_{3}$ are trade-off parameters that weight the importance of the corresponding terms. And our optimization objective is to learn a target model $\mathcal{G}$ according to:
\begin{equation}
    \mathcal{G}=\mathop{\arg}\mathop{\min}_{G_{sem}} \mathop{\min}_{\substack{G_{sem}\\G_{eg}}} \mathop{\max}_{\substack{D_{sem}\\D_{eg}}} \mathcal{L}
\label{eq:7}
\end{equation}
\par The full training procedure of our method consists of three steps: 1) jointly optimizing $G_{sem}$, $D_{sem}$, $G_{eg}$, and $D_{eg}$ by semantic segmentation loss, edge loss, and two adversarial learning loss over the source and target domains; 2) generating pseudo labels and correspond entropy maps by $G_{sem}$ over the target domain; 3) optimizing $G_{sem}$ by uncertainty-adaptive self-supervised loss. 

\begin{table*}
    \caption{Evaluation results of semantic segmentation by adapting from SYNTHIA to Cityscapes. The mechanism “T”, “A”, and “S” mean image translation, adversarial learning, and self-supervised learning, respectively. We show the mIoU ($\%$) of the 13 classes (mIoU*) excluding classes with “*”. “-” represents the method does not report the corresponding experimental result. The best results are highlighted in \textbf{bold}.}
    \setlength\tabcolsep{3.35pt}
    \centering
    \begin{tabular}{l | c | c c c c c c c c c c c c c c c  c | c | c}
    \toprule
    \multicolumn{20}{c}{SYNTHIA$\rightarrow$Cityscapes}\\
    \hline
    Methods & \rotatebox{90}{Mech.} &\rotatebox{90}{road} & \rotatebox{90}{sidewalk } & \rotatebox{90}{building} & \rotatebox{90}{wall*} & \rotatebox{90}{fence*} & \rotatebox{90}{pole*} & \rotatebox{90}{light} & \rotatebox{90}{sign} & \rotatebox{90}{veg.}  & \rotatebox{90}{sky} & \rotatebox{90}{person} & \rotatebox{90}{rider} & \rotatebox{90}{car}  & \rotatebox{90}{bus} &  \rotatebox{90}{mbike} & \rotatebox{90}{bike} & mIoU & mIoU* \\
    \hline
    AdaSegNet \cite{tsai2018learning}& A &	81.7&	39.1&	78.4&	11.1&	0.3&	25.8&	6.8&	9.0&	79.1&	80.8&	54.8&	21.0&	66.8&	34.7&	13.8&	29.9&	39.6&	45.8\\
    ADVENT \cite{vu2019advent} & A &	85.6&	42.2&	79.7&	8.7&	0.4&	25.9&	5.4&	8.1&	80.4&	84.1&	57.9&	23.8&	73.3&	36.4&	14.2&	33.0&	41.2&	48.0\\
    CLAN \cite{luo2019taking}&	A& 81.3&	37.0&	80.1&	-&	-&	-&	16.1&	13.7&	78.2&	81.5&	53.4&	21.2&	73.0&	32.9&	22.6&	30.7&	-&	47.8\\
    AdaptPatch \cite{tsai2019domain} &A&	82.4&	38.0&	78.6&	8.7&	0.6&	26.0&	3.9&	11.1&	75.5&	84.6&	53.5&	21.6&	71.4&	32.6&	19.3&	31.7&	40.0&	46.5\\
    % SIBAN \cite{luo2019significance}&A&	82.5&	24.0&	79.4&	-&	-&	-&	16.5&	12.7&	79.2&	82.8&	58.3&	18.0&	79.3&	25.3&	17.6&	25.9&	46.3&	-\\
    PyCDA \cite{lian2019constructing}&S&75.5& 30.9 &	\textbf{83.3}& \textbf{20.8}& 0.7& 32.7& \textbf{27.3} &\textbf{33.5} &84.7& 85.0 &\textbf{64.1}& 25.4 &85.0& 45.2 &21.2& 32.0& 46.7& 53.3\\
    CCM \cite{li2020content} &S& 79.6& 36.4 &80.6& 13.3 &0.3 &25.5& 22.4& 14.9 &81.8& 77.4 &56.8 &25.9 &80.7 &45.3 &29.9 &\textbf{52.0} &45.2&52.9 \\
    \hline
    FDA \cite{yang2020fda}&TS&	79.3&	35.0&	73.2&	-&	-&	-&	19.9&	24.0&	61.7&	82.6&	61.4&	\textbf{31.1}&	83.9&	40.8&	\textbf{38.4}&	51.1&	-&	52.5\\
    IntraDA \cite{pan2020unsupervised}&AS&	84.3&	37.7&	79.5&	5.3&	0.4&	24.9&	9.2&	8.4&	80.0&	84.1&	57.2&	23.0&	78.0&	38.1&	20.3&	36.5&	41.7&	48.9\\
    FADA \cite{wang2020classes}&AS&	84.5&	40.1&	83.1 &	4.8&	0.0 &	\textbf{34.3} &	20.1&	27.2&	\textbf{84.8} &	84.0&	53.5&	22.6&	85.4 &	43.7&	26.8&	27.8&	45.2&	52.5\\
    DAST \cite{yu2021dast} &AS&	87.1&	44.5&	82.3 &	10.7 &	\textbf{0.8}&	29.9&	13.9&	13.1&	81.6&	\textbf{86.0}&	60.3&	25.1&	83.1&	40.1&	24.4&	40.5&	45.2&	52.5\\
    \hline
    BDL \cite{li2019bidirectional} & TAS& 86.0&  46.7 & 80.3 & -&  -&  -&  14.1&  11.6&  79.2&  81.3&  54.1&  27.9&  73.7&  42.2&  25.7&  45.3 &- &  51.4\\
    TIR \cite{yang2020fda}&TAS&92.6& 53.2& 79.2& -&-&-&1.6& 7.5& 78.6& 84.4& 52.6& 20.0& 82.1& 34.8& 14.6& 39.4& - & 49.3 \\
    LDR \cite{yang2020label}&TAS &85.1& 44.5& 81.0& - & - & - &16.4 &15.2 &80.1 &84.8 &59.4 &31.9& 73.2& 41.0& 32.6 &44.7&- & 53.1\\
    UDACT \cite{lee2020unsupervised} &TAS&	\textbf{93.3}&	\textbf{54.0}&	81.3&	14.3&	0.7&	28.8&	21.3&	22.8&	82.6&	83.3&	57.7&	22.8&	83.4&	30.7&	20.2&	47.2&	46.5&	53.9\\
    \hline        
    % SourceOnly \cite{tsai2019domain}& - &	55.6&	22.7&	68.6&	4.3&	0.1&	23.0&	5.6&	9.1&	77.2&	75.9&	54.7&	8.7&	81.5&	23.9&	8.4&	8.8&	38.5&	33.0 \\
    SourceOnly & - &	55.9&	22.7&	72.1&	9.3&	0.1&	24.7&	10.3&	10.4&	73.8&	77.9&	54.9&	20.5&	41.2&	31.7&	8.3&	11.5&	32.8&	37.8 \\
    % Baseline &A&	77.3&	36.3&	79.0&	10.5&	0.3&	26.4&	12.3&	11.4&	79.3&	82.7&	55.8&	21.7&	63.8&	37.6&	20.5&	32.2&	46.9&	40.5\\
    % Ours &AS&	91.5&	51.2&	82.0&	\textbf{20.0}&	\textbf{1.2}&	30.6&	11.2&	15.4&	82.6&	85.2&	59.4&	24.3&	84.8&	45.0&	26.8&	48.5&	\textbf{54.5}&	\textbf{47.5}\\  
    Ours &AS&	92.0&	53.9&	82.0&	10.1&	0.2&	32.8&	13.3&	26.0&	83.6&	84.4&	63.0&	21.4&	\textbf{86.7}&	\textbf{46.8}&	24.7&	49.0&	\textbf{48.1}&	\textbf{55.9}\\  
    \toprule
    \end{tabular}
    \label{tab:2}
\end{table*}

\section{Experiments}
\par In this section, following the common protocol of previous works \cite{hoffman2016fcns,luo2019taking,vu2019advent}, we conduct experiments on the two-challenging synthetic-to-real unsupervised domain adaptation tasks, i.e., GTAV$\rightarrow$Cityscapes, and SYNTHIA$\rightarrow$Cityscapes. Specifically, we use GTAV or SYNTHIA datasets with pixel-level annotations as the source domain and Cityscapes dataset without any annotations as the target domain.

\subsection{Datasets}
\par \textbf{Cityscapes} is a real-world urban scene image dataset, which provides 3975 images, each of which has a resolution of 2048$\times$1024, collected from 50 cities in Germany \cite{cordts2016cityscapes}. Following the standard protocols \cite{hoffman2016fcns,luo2019taking,vu2019advent}, we use the 2975 images from Cityscapes dataset training set as the unlabeled target domain for training, and evaluate our method on the 500 images from the validation set. 

\par \textbf{GTAV} is a large synthetic dataset containing 24966 high quality labeled urban scene images with a resolution of 1914$\times$1052 from open-world computer games, Grand Theft Auto V \cite{richter2016playing}. The 19 compatible semantic classes between GTAV and Cityscapes are selected in the experiment.

\par \textbf{SYNTHIA} is another synthetic urban scene dataset \cite{ros2016synthia}. Following previous works, we use the SUNTHIA-RAND-CITYSCAPES subset that contains 9400 annotated images with a resolution of 1280$\times$760 and shares 16 semantic classes with Cityscapes. In the training stage, we consider the 16 common classes with the Cityscapes. In the evaluation stage, 16- and 13-class subsets are used to make quantitative assessment. 

\subsection{Implementation Details}
\par All the experiments in this paper \footnote{The source code will be made publicly
available.} are implemented with Pytorch in a single NVIDIA GTX 1080Ti GPU. Limited by the GPU memory, during the training stage, the resolution of Cityscapes images is resized to 1024$\times$512, and that of GTAV images is resized to 1280$\times$720. The resolution of SYNTHIA images remains unchanged. For the sake of fair comparison, like most of the state-of-the-art methods, we do not use any data augment technique. 

\par For the semantic stream $G_{sem}$, following most of the state-of-the-art methods, we use ResNet-101 architecture \cite{he2016deep} with pretrained parameters from ImageNet \cite{deng2009imagenet}. The implementation of edge stream $G_{eg}$ follows the work \cite{takikawa2019gated}. $G_{eg}$ is mainly composed of three residual blocks, and each block is followed by a gated convolutional layer, ensuring $G_{eg}$ only processes edge related information. For the discriminator $D_{sem}$, we apply the same architecture used in \cite{vu2019advent}. For the discriminator $D_{eg}$, we adopt a simple structure consisting of three 4$\times$4 convolutional layers with a stride of 2, and one 1$\times$1 convolutional layer. Except for the last layer, each convolutional layer is followed by a Leaky-ReLU with a slope of 0.2.

\par To verify the robustness of our method, the hyper-parameters keep the same in both tasks. For our joint loss, the values of $\lambda_{1}$ to $\lambda_{3}$ are set as 1$e^{-3}$, 20, and 1$e^{-3}$, respectively. The weight factor $\alpha$ in entropy reweighting adversarial loss is set to 10. In terms of the whole architecture training, the SGD optimizer with a learning rate of 2.5$e^{-4}$, momentum of 0.9, and a weight decay of 5$e^{-4}$ is utilized to train $G_{sem}$ and $G_{eg}$. Two Adam optimizers with a learning rate of 1$e^{-4}$ are used for training $D_{sem}$ and $D_{eg}$, respectively.

\begin{figure*}[!t]
    \centering
    \subfloat[]{
      \includegraphics[width=1.6in]{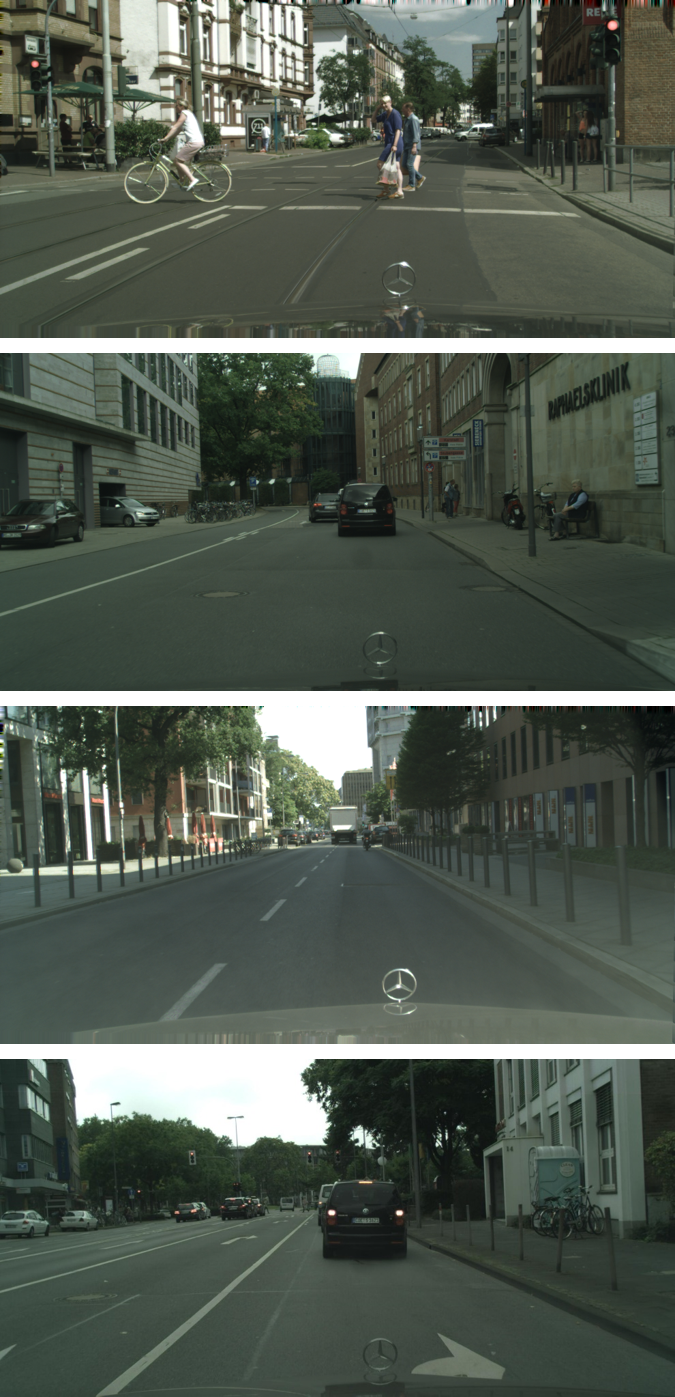}
    \label{fig:CSCD}}
    \centering
    \subfloat[]{
      \includegraphics[width=1.6in]{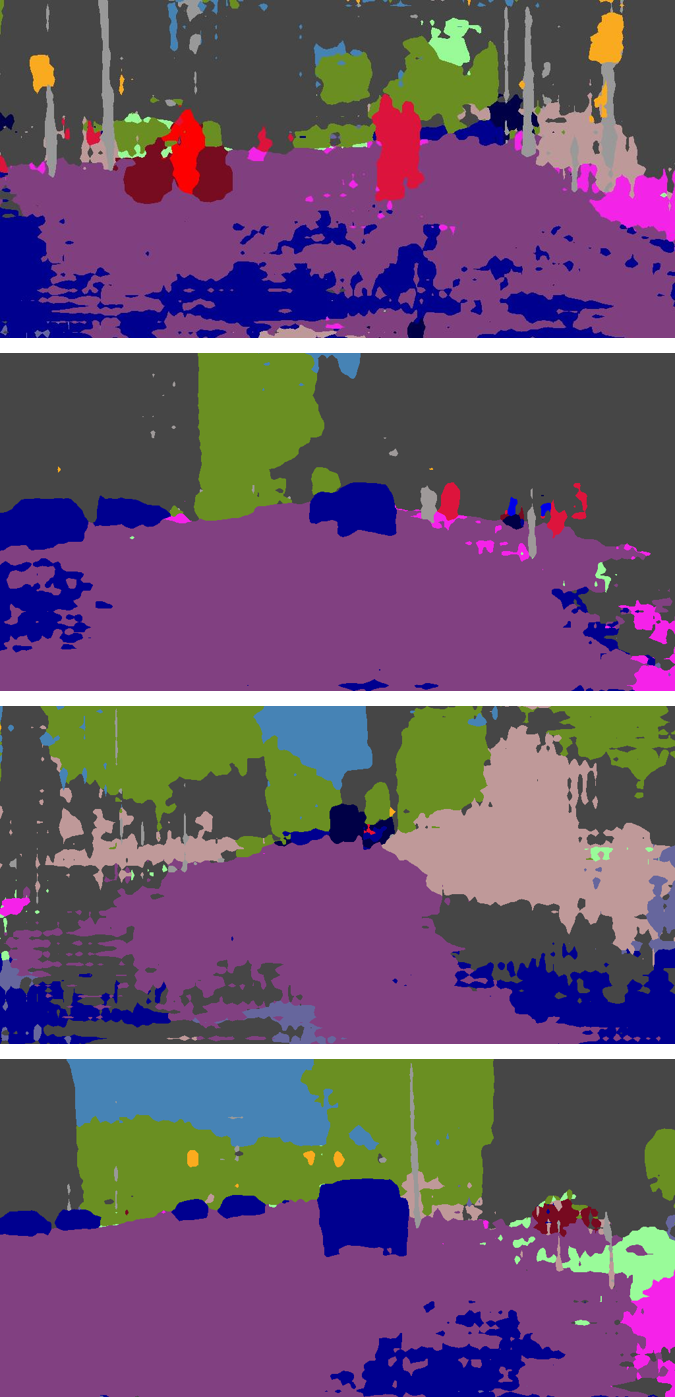}
    \label{fig:CSCD}}
    \centering
    \subfloat[]{
      \includegraphics[width=1.6in]{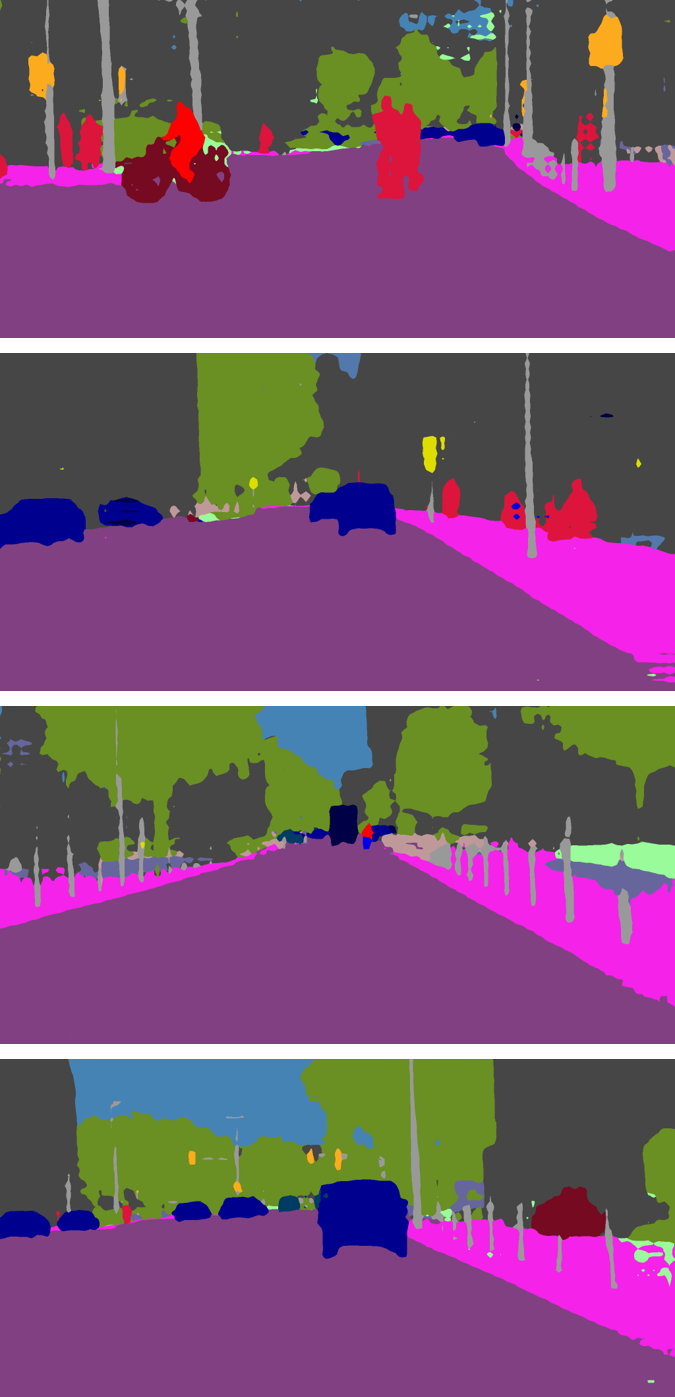}
    \label{fig:CSCD}}
    \centering
    \subfloat[]{
      \includegraphics[width=1.6in]{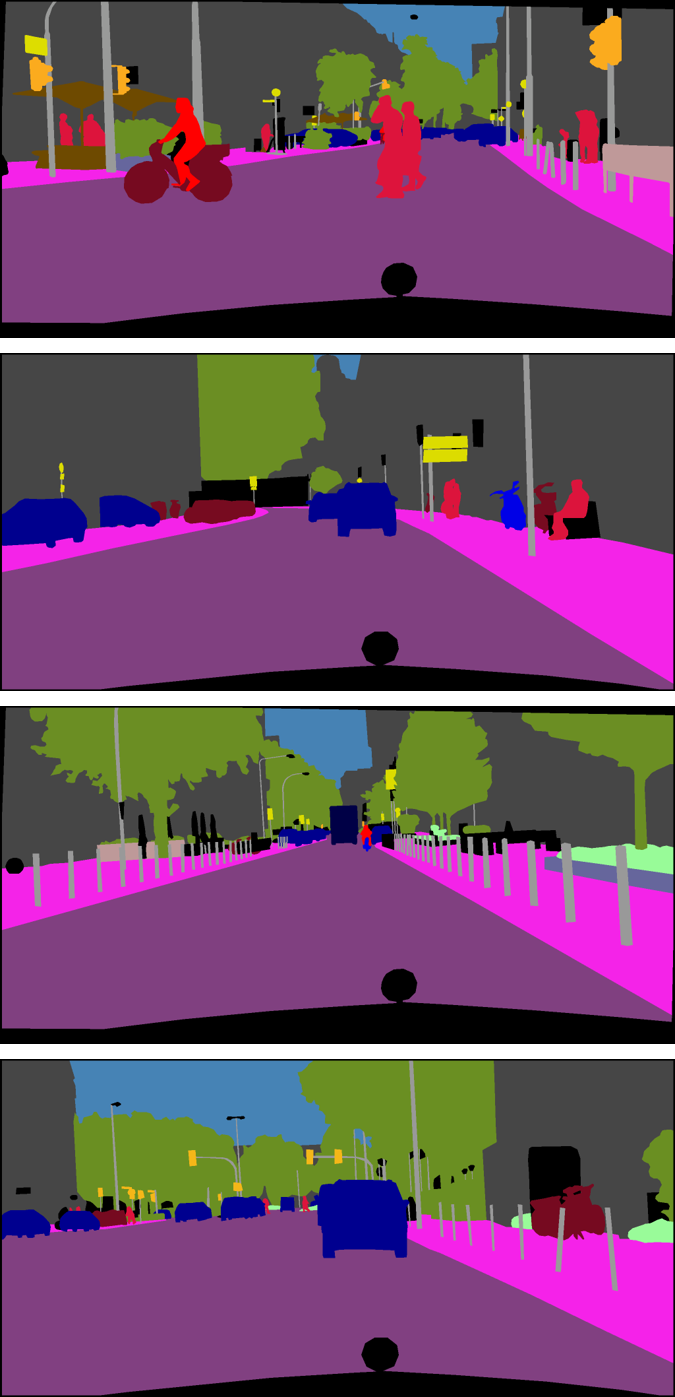}
    \label{fig:CSCD}}
    \caption{Qualitative results on the GTAV$\rightarrow$Cityscapes task. (a) Input images from Cityscapes. (b) Segmentation results without domain adaptation. (c) Segmentation results of the proposed method.(d) Ground truth.}
    \label{fig:3}
\end{figure*}

\begin{figure}[!t]
    \centering
    \subfloat[]{
      \includegraphics[width=0.95in]{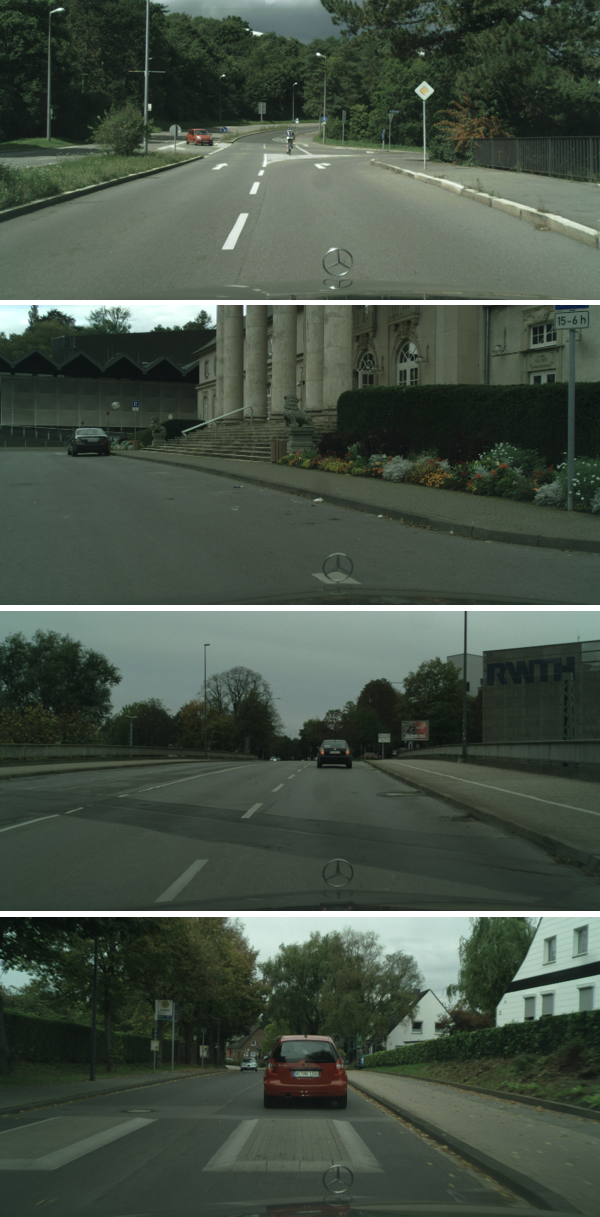}
    \label{fig:CSCD}}
    \subfloat[]{
      \includegraphics[width=0.95in]{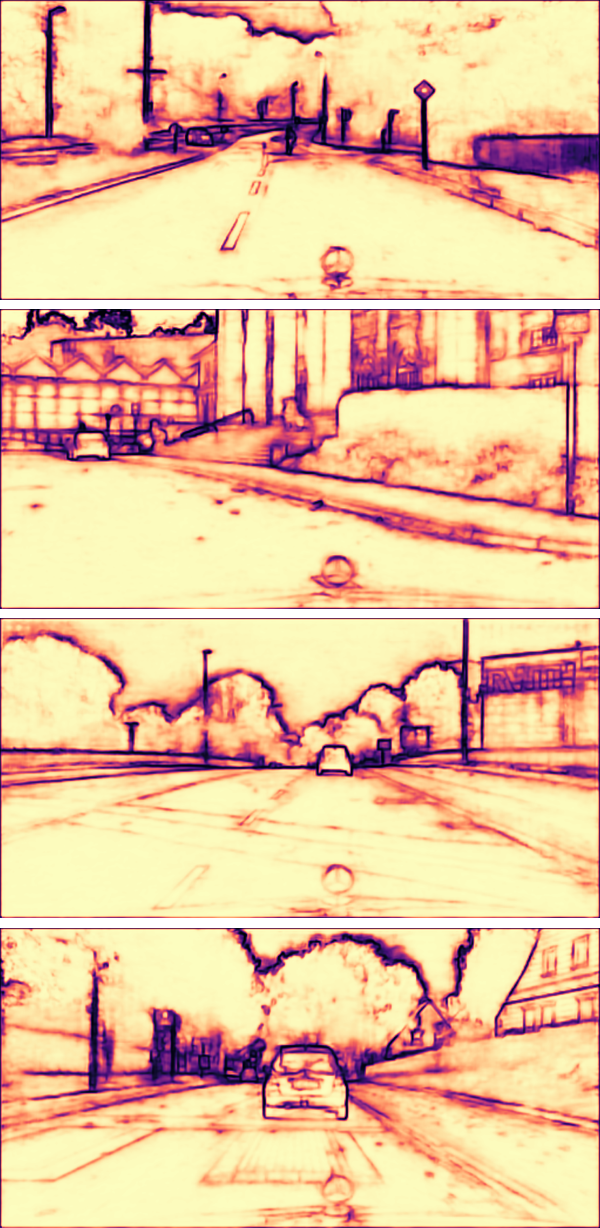}
    \label{fig:CSCD}}
    \subfloat[]{
      \includegraphics[width=0.95in]{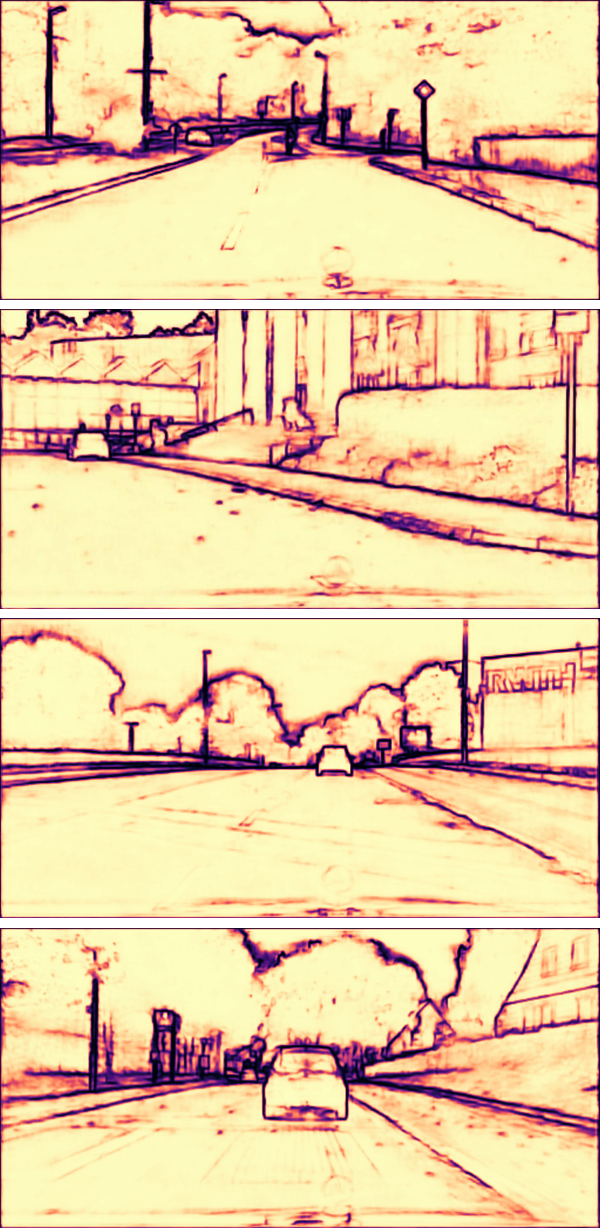}
    \label{fig:CSCD}}
    \caption{Illustrations of produced boundary maps. (a) Input images from Cityscapes. (b) Semantic boundary maps without adversarial learning. (c) Semantic boundary maps with adversarial learning.}
    \label{fig:4}
\end{figure}
\begin{figure}[!t]
    \centering
    \subfloat[]{
      \includegraphics[width=1.43in]{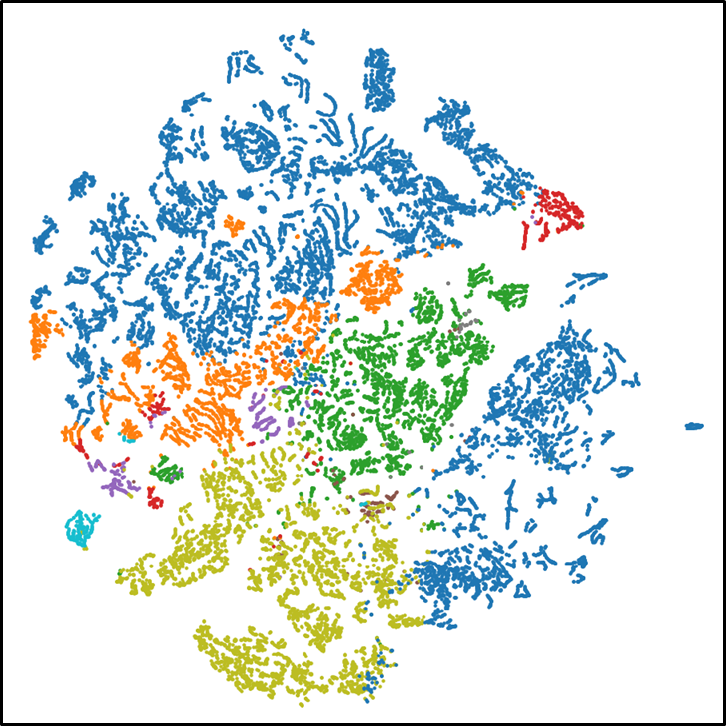}
    \label{fig:CSCD}}
    \centering
    \subfloat[]{
      \includegraphics[width=1.43in]{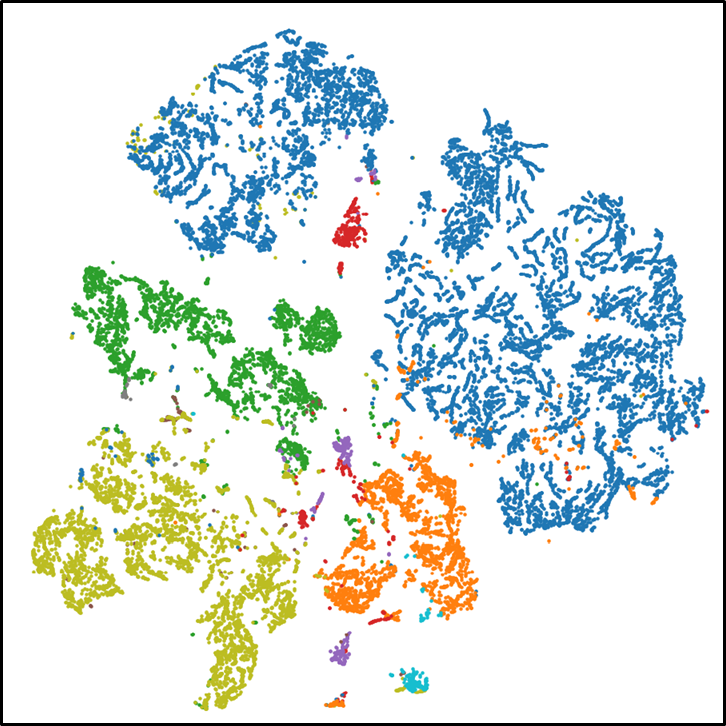}
    \label{fig:CSCD}}
    \caption{The t-SNE visualization of embedded semantic features over the target domain. (a) Features without the guide of edge information. (b) Features with the guide of edge information.}
    \label{fig:5}
\end{figure}
\subsection{Performance Comparison}

% \par In this subsection, we compare our proposed method with the existing state-of-the-art methods, from adversarial learning to self-supervised learning, and some hybrid methods, including AdaSegNet \cite{tsai2018learning}, ADVENT  \cite{vu2019advent}, CLAN \cite{luo2019taking}, BDL \cite{li2019bidirectional}, AdaptPatch \cite{tsai2019domain}, PyCDA \cite{lian2019constructing}, IntraDA \cite{pan2020unsupervised}, TIR \cite{kim2020learning}, FDA	\cite{yang2020fda}, FADA \cite{wang2020classes}, CCM \cite{wang2020classes}, LDR \cite{yang2020label} \cite{li2020content}, DAST \cite{yu2021dast}, and UDACT \cite{lee2020unsupervised}. For the sake of fairness, all the reported methods use ResNet-101 as the backbone network and data augment technique is not used. The per-class Intersection-Over-Union (IoU) and mean IoU (mIoU) are adopted as the evaluation criteria.

\par In this subsection, we compare our proposed method with the existing state-of-the-art methods \cite{tsai2018learning,vu2019advent,luo2019taking,li2019bidirectional,tsai2019domain,lian2019constructing,pan2020unsupervised,kim2020learning,yang2020fda,wang2020classes,yang2020label,li2020content,yu2021dast,lee2020unsupervised}. For the sake of fairness, all the reported methods use ResNet-101 as the backbone network and the data augment technique is not used. The per-class Intersection-Over-Union (IoU) and mean IoU (mIoU) are adopted as the evaluation criteria.

\par \textbf{GTAV to Cityscapes.} Table \ref{tab:1} displays the comparison results from GTAV to Cityscapes. First, all domain adaptation methods outperform the model without domain adaptation (SourceOnly) by large performance margins. Then, our method has the best mIoU 52.8$\%$, which is significantly better than that of the compared state-of-the-art methods. Compared with some adversarial training and self-supervised learning methods, such as CLAN and PyCDA, our method improves by 9.6$\%$ and 5.4$\%$ mIoU and has significant gains in almost all classes. Moreover, some methods also combine the adversarial learning or image translation with self-supervised learning and achieve decent performance, like FDA, DAST, and UDACT. Compared to these methods, our approach still has a significant improvement. Figure \ref{fig:3} presents some qualitative results \footnote{For more qualitative results, please see the supplementary materials.} produced by our methods. 
% As is evident, all domain adaptation methods outperform the models without domain adaptation (SourceOnly) by a large performance margin. Moreover, the performance of our proposed method is significantly better than that of the compared state-of-the-art methods. Figure \ref{fig:3} presents some qualitative results. We could see that the segmentation maps generated by our model are close to the ground truth and have high intra-class compactness and precise boundaries.

\par \textbf{SYNTHIA to Cityscapes.} Table \ref{tab:2} reports the comparison results on the SYNTHIA$\rightarrow$Cityscapes task. Like the previous work, we also report two mIoU metrics: 13 classes of mIoU* and 16 classes of mIoU. According to Table \ref{tab:2}, it is obvious that our proposed method can outperform other state-of-the-art methods on both 13-class and 16-class with mIoU of 55.9$\%$ and 48.1$\%$. In summary, these results obtained on both tasks reveal the effectiveness and superiority of our architecture in learning domain-invariant representations for UDA in semantic segmentation.

\subsection{Discussion}
\par We further report the ablation study results to demonstrate the performance contribution of each element in our proposed method in Table \ref{tab:3}. It can be seen that each element contributes to the final success of the adaptation. The proposed method outperforms the “Baseline” by +7.7$\%$ and +6.9$\%$ with GTAV and SYNTHIA as the source, respectively.

\begin{table}[!t]
        \caption{Ablation study of the proposed method in terms of mIoU ($\%$) on the two tasks. Here, “Baseline” represents the original adversarial learning without entropy reweighting.}
        \centering
        \setlength\tabcolsep{3.4pt}
        \begin{tabular}{ccccc|cc}
        \toprule
        Baseline &   $\mathcal{L}_{sem}^{adv}$ & $\mathcal{L}_{eg}^{con}$ & $\mathcal{L}_{eg}^{adv}$ & $\mathcal{L}_{uasl}$ & GTAV&	SYNTHIA \\
        \hline
        $\checkmark$ &  & & & & 45.1&	41.2\\
        $\checkmark$  & $\checkmark$ &  & & & 46.3&	41.8\\
        $\checkmark$ & $\checkmark$  & $\checkmark$ & & &	48.4&	43.7\\
        % $\checkmark$ & $\checkmark$  &  & & $\checkmark$  &	48.9&	44.0\\
         $\checkmark$ & $\checkmark$ & $\checkmark$ & $\checkmark$ &   &	49.2&	44.4\\
        $\checkmark$ & $\checkmark$& $\checkmark$&$\checkmark$ &$\checkmark$ &	\textbf{52.8}&	\textbf{48.1}\\
        \toprule
        \end{tabular}
        \label{tab:3}
\end{table}

% \begin{table*}[!t]
%         \caption{Ablation study of the proposed method in terms of mIoU ($\%$). Here, $\hat{\mathcal{L}}_{sem}^{adv}$ represents the original adversarial learning loss without entropy reweighting, and $\hat{\mathcal{L}}_{eg}^{con}$ denotes the combination of $\mathcal{L}_{eg}^{seg}$ and $\mathcal{L}_{eg}^{con}$.}
%         \centering
%         \begin{tabular}{l|cccccc|cc}
%         \toprule
%         & $\mathcal{L}_{sem}^{seg}$ & $\hat{\mathcal{L}}_{sem}^{adv}$ & $\mathcal{L}_{sem}^{lov}$ & $\mathcal{L}_{sem}^{adv}$ & $\hat{\mathcal{L}}_{eg}^{con}$ & $\mathcal{L}_{eg}^{adv}$ & $\mathcal{L}_{uasl}$ & GTAV&	SYNTHIA \\
%         \hline
%         SourceOnly & $\checkmark$ & & & & & & &	36.5&	32.8\\
%         Baseline & $\checkmark$ & $\checkmark$ & & & & & &	43.9&	40.5\\
%         & $\checkmark$ & $\checkmark$ & $\checkmark$ & & & & & 45.1&	41.2\\
%         & $\checkmark$ & & $\checkmark$  & $\checkmark$ & & & & 46.3&	41.8\\
%         & $\checkmark$ & & $\checkmark$ &$\checkmark$  & $\checkmark$ & & &	48.4&	43.7\\
%         & $\checkmark$ & & $\checkmark$ & $\checkmark$ & $\checkmark$ & $\checkmark$ &  &	49.2&	44.4\\
%         Ours & $\checkmark$ & & $\checkmark$ & $\checkmark$& $\checkmark$&$\checkmark$ &$\checkmark$ &	\textbf{52.8}&	\textbf{48.1}\\
%         \toprule
%         \end{tabular}
%         \label{tab:3}
% \end{table*}
\begin{table}
    \caption{Parameter analysis of the weighting factor $\alpha$ for the entropy reweighting adversarial loss in terms of mIoU ($\%$).}
    \centering
    \begin{tabular}{c c c c c c c c}
    \toprule
    \multicolumn{8}{c}{GTAV$\rightarrow$Cityscapes}\\
    \hline
    $\alpha$ & 0 & 1 & 5 & 10& 15 & 20 & 30  \\
    \hline
    mIoU & 45.1 & 45.2 & 45.8 & \textbf{46.3} & 46.0 & 45.3 & 43.0 \\  
    \toprule
    \end{tabular}
    \label{tab:4}
\end{table}

\begin{table}
    \caption{Comparison of the proposed UASL and standard self-supervised learning method with different thresholds $T$ in terms of mIoU ($\%$).}
    \centering
    \begin{tabular}{l|l|cc}
    \toprule
    \multicolumn{2}{c|}{SL Type}  & GTAV & SYNTHIA \\
    \hline
    \multirow{3}{*}{SL} & $T=0$  & 51.3   & 47.1   \\
    ~ & $T=0.5$     & 52.2  & 47.6     \\
    ~ & $T=0.9$     & 52.5 &  47.4  \\
    \hline
    \multicolumn{2}{c|}{UASL}  & \textbf{52.8}  & \textbf{48.1}     \\
    \toprule
    \end{tabular}
    
    \label{tab:5}
\end{table}
\par Specifically, our proposed image-level entropy reweighting methods can enhance the model adaptation ability in the hard samples, thereby improving performance by 1.2$\%$ and 0.6$\%$. Besides, weight factor $\alpha$ is an important parameter for the entropy reweighting adversarial loss, which controls the extra adaptation degree to hard samples, we evaluate the performance of the semantic stream with different $\alpha$ on the GTAV$\rightarrow$Cityscapes task, as shown in Table 4. When $\alpha$=0, the loss is equal to the standard adversarial loss. As $\alpha$ increases, the loss of high-entropy samples (i.e., hard samples) is enlarged and hard samples could get better adaptation. However, if $\alpha$ is too large, the network could merely focus on the adaptation of hard samples, which may damage the adaptation performance in easy samples. 

\par Subsequently, utilizing low-level edge information can further enhance adaptation performance and lead to a large performance boost. Since the small domain gap in low-level edge features can be narrowed to some extent by semantic adversarial learning, using an independent edge stream to generate semantic boundaries and align the target prediction maps with them can improve mIoU by 2.1$\%$ and 1.9$\%$. After explicitly performing adversarial learning for edge information, the mIoU get further improved by 0.8$\%$ and 0.7$\%$. This obvious performance improvement (+2.9$\%$ and +2.6$\%$) clearly demonstrates our argument. Figure \ref{fig:4} illustrates some produced semantic boundary maps. Obviously, through adversarial learning, the small inter-domain gap in edge features get further narrowed, thereby producing more accurate semantic boundaries. Then, in Figure \ref{fig:5}, we additionally give a contrastive analysis between semantic feature distributions without and with the guide of edge information by t-SNE \cite{van2008visualizing}, which reveals that our edge consistency loss can enforce the alignment of semantic features, leading to clearer and more discriminative clusters over the target domain. 

\par
\begin{figure}[!t]
    \centering
    \includegraphics[scale=0.35]{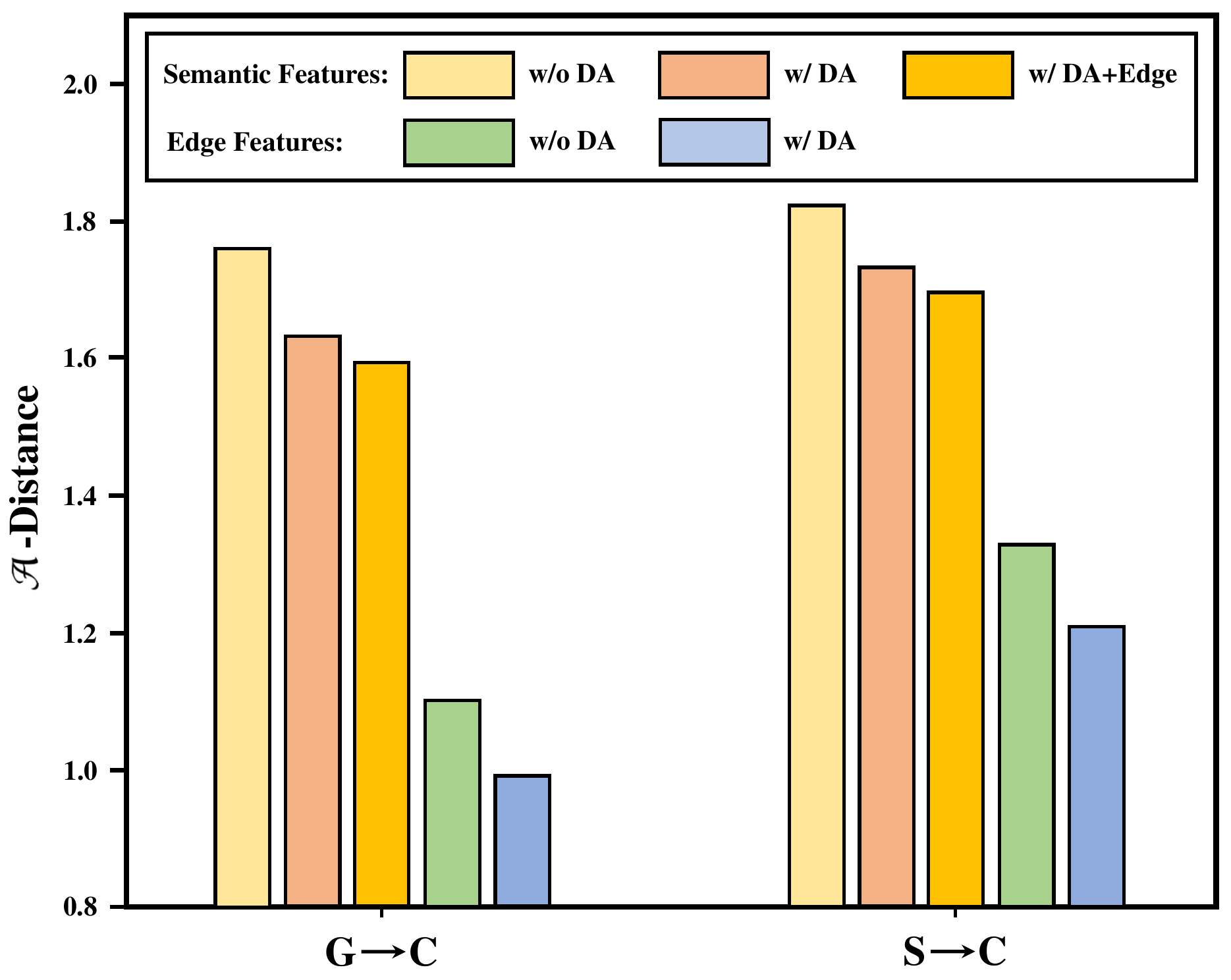}
    \caption{$\mathcal{A}$-distance of semantic and edge feature representations on the two tasks.}
    \label{fig:6}
\end{figure}

 To further verify our argument, we display $\mathcal{A}$-distance \cite{ben2007analysis} of semantic features and edge features in Figure \ref{fig:6}, which is a commonly used metric for measuring domain discrepancy. Since it is difficult to compute the exact $\mathcal{A}$-distance, a proxy distance is defined as $\hat{d}_{\mathcal{A}}=2(1-2 \epsilon)$ \cite{ben2007analysis}, where $\epsilon$ is the generalization error of a classifier (SVM in this paper) trained on the binary problem of distinguishing samples between the source and target domains. From Figure \ref{fig:6}, we could see that $\hat{d}_{\mathcal{A}}$ on edge features is obviously smaller than $\hat{d}_{\mathcal{A}}$ on semantic features, which proves that low-level edge information has a smaller inter-domain gap. After performing semantic adversarial learning, the $\hat{d}_{\mathcal{A}}$ on semantic features is reduced. Moreover, the $\hat{d}_{\mathcal{A}}$ on semantic features become smaller through the guide of edge information, which verifies that edge information can be explicitly used to facilitate the transfer of semantic information. 

\par Lastly, through UASL, the mIoU performance of the proposed method reaches 52.8$\%$ and 48.1$\%$. In addition, we also compare UASL to the standard SL method with three commonly used thresholds: $T=0$ \cite{pan2020unsupervised}, $T=0.5$  \cite{lian2019constructing}, and $T=0.9$ \cite{kim2020learning}. The relevant results are reported in Table \ref{tab:5}. We could see that the threshold of generating pseudo-label  significantly affects the performance of standard SL. And different domain adaptation tasks may have diverse optimal thresholds. It is difficult to choose a suitable threshold. In contrast, our UASL does not need to select a threshold. It can fully utilize target label information, adaptively pay large weights to high-confident pixels and suppress the effects of low-confident pixels, thereby obtaining better performance.

\section{Conclusion}
\par  In this paper, we present a new domain adaptation approach by leveraging the low-level edge information that is easy to adapt to guide the transfer of high-level semantic information. Specifically, we propose a semantic-edge domain adaptation architecture. The semantic stream adopts the existing entropy adversarial learning approach. To better adapt hard target samples, an entropy reweighting method is presented to make the network pay more attention to hard samples. The edge stream can produce semantic boundaries. To make the target predicted boundaries more precise, adversarial learning is performed on the edge stream. For the purpose of explicitly guiding the transfer of semantic features, an edge consistency loss function is presented to ensure the consistency between the target semantic map and boundary map. Lastly, an uncertainty-adaptive self-supervised learning is proposed to further fit the distribution of the target domain. The experimental results in the two UDA segmentation scenarios from synthetic to real demonstrate that our method obtains better results than the existing state-of-the-art works.

% \begin{acks}
% Thanks xxxxx
% \end{acks}

\bibliographystyle{ACM-Reference-Format}
\bibliography{sample-base}


\end{document}